\documentclass[12pt,letterpaper]{article}
\usepackage[a4paper, total={7in, 10in}]{geometry}

\usepackage{graphicx}
\usepackage{helvet}
\usepackage{authblk}
\usepackage{hyperref}
\usepackage{amsmath} 
\usepackage{amssymb} 
\usepackage{orcidlink} 
\usepackage[super,comma,sort&compress]
   {natbib}

\usepackage{setspace}
\AtBeginDocument{\doublespacing}

\usepackage{authblk}
\usepackage{url}           
\usepackage{booktabs}       
\usepackage{amsfonts}       
\usepackage{nicefrac}      
\usepackage{microtype}      
\usepackage{lipsum}
\usepackage{fancyhdr}       
\usepackage{graphicx}       
\usepackage{xcolor}         
\graphicspath{{media/}}     
\usepackage{float}
\usepackage{amsmath,amsfonts}
\usepackage{algorithmic}
\usepackage{algorithm}
\usepackage{array}
\usepackage{textcomp}
\usepackage{stfloats}
\usepackage{url}
\usepackage{mathtools}
\usepackage{mathrsfs}
\usepackage{amsthm}
\usepackage{amssymb}
\usepackage{multirow}
\usepackage{authblk}
\usepackage{footmisc}
\usepackage{enumitem}
\usepackage[normalem]{ulem}
\usepackage{dsfont}
\usepackage{subcaption}
\usepackage{tabularx}
\usepackage{makecell}

\newcolumntype{L}{>{\raggedright\arraybackslash}X}   
\newcolumntype{Y}{>{\centering\arraybackslash}X}

\def\loss{\mathrm{loss}}

\def\*#1{\boldsymbol{#1}}
\makeatletter
\newcommand{\vx}{\vec{x}\@ifnextchar{^}{\,}{}}
\makeatother

\newtheorem{remark}{Remark}

\definecolor{DarkBlue}{rgb}{0,0.28,0.67}

\makeatletter
\renewcommand{\maketitle}{\bgroup\setlength{\parindent}{0pt}
\begin{flushleft}
  \textbf{\@title}
  
  \@author
\end{flushleft}\egroup}
\makeatother

\usepackage{tabularray}
\UseTblrLibrary{booktabs}

\usepackage{seqsplit}

\title{Training Together, Diagnosing Better: Federated Learning for Collagen VI–Related Dystrophies}
\date{}

\author[1]{Astrid Brull}
\author[2]{Sara Aguti}
\author[1]{Véronique Bolduc}
\author[1]{Ying Hu}
\author[3]{Daniel M. Jimenez-Gutierrez}
\author[3]{Enrique Zuazua}
\author[3,10]{Joaquin Del-Rio}
\author[3]{Oleksii Sliusarenko}
\author[5,7]{Haiyan Zhou}
\author[2,6,7]{Francesco Muntoni}
\author[1,9]{Carsten G. Bönnemann}
\author[3]{Xabi Uribe-Etxebarria}

\affil[1]{Neurogenetics and Neuromuscular Disorders of Childhood Section, National Institute of Neurological Disorders and Stroke, National Institutes of Health, Bethesda, MD 20892, USA}
\affil[2]{Neurodegenerative Disease Department, UCL Queen Square Institute of Neurology, University College London, London WC1N 3BG, UK}
\affil[3]{Sherpa.ai, Erandio, Bizkaia 48950, Spain}
\affil[5]{Genetics and Genomic Medicine Research and Teaching Department, Great Ormond Street Institute of Child Health, University College London, London WC1N 1EH, UK}
\affil[6]{Developmental Neurosciences Research and Teaching Department, Great Ormond Street Institute of Child Health, University College London, London WC1N 1EH, UK}
\affil[7]{NIHR Great Ormond Street Hospital Biomedical Research Centre, London WC1N 1EH, UK}
\affil[9]{Corresponding author at: Neurogenetics and Neuromuscular Disorders of Childhood Section, National Institute of Neurological Disorders and Stroke, National Institutes of Health, Bethesda, MD 20892, USA (Email: carsten.bonnemann@nih.gov)}
\affil[10]{Corresponding author at: Sherpa.ai, Erandio, Bizkaia 48950, Spain (Email: research@sherpa.ai)}

\begin{document}

\maketitle

\section*{Abstract}

\begin{figure}[hbtp]
    \centering
    \includegraphics[width=0.72\linewidth]{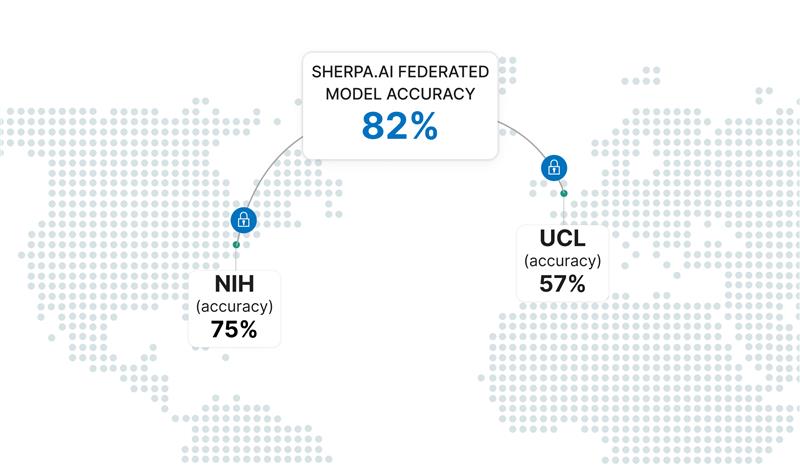}
    \caption{Overview of the Sherpa.ai FL setup showing individual institutional accuracies and the resulting global federated model performance.}
    \label{fig:world_accuracy_comparison}
\end{figure}

The application of Machine Learning (ML) to the diagnosis of rare diseases, such as collagen VI-related dystrophies (COL6-RD), is fundamentally limited by the scarcity and fragmentation of available data. Attempts to expand sampling across hospitals, institutions, or countries with differing regulations face severe privacy, regulatory, and logistical obstacles that are often difficult to overcome. The Federated Learning (FL) provides a promising solution by enabling collaborative model training across decentralized datasets while keeping patient data local and private. Here, we report a novel global FL initiative using the Sherpa.ai FL platform, which leverages FL across distributed datasets in two international organizations for the diagnosis of COL6-RD, using collagen VI immunofluorescence microscopy images from patient-derived fibroblast cultures. Our solution resulted in an ML model capable of classifying collagen VI patient images into the three primary pathogenic mechanism groups associated with COL6-RD: exon skipping, glycine substitution, and pseudoexon insertion. This new approach achieved an F1-score of 0.82, outperforming single-organization models (0.57–0.75). These results demonstrate that FL substantially improves diagnostic utility and generalizability compared to isolated institutional models (see Figure~\ref{fig:world_accuracy_comparison}). Beyond enabling more accurate diagnosis, we anticipate that this approach will support the interpretation of variants of uncertain significance and guide the prioritization of sequencing strategies to identify novel pathogenic variants.



\section{Introduction}
Collagen VI is an extracellular matrix (ECM) protein that forms a beaded microfibrillar network surrounding skeletal muscle fibers, contributing to muscle function and myofiber integrity. It is composed of three alpha chains, $\alpha$1(VI), $\alpha$2(VI) and $\alpha$3(VI), encoded by \textit{COL6A1, COL6A2 and COL6A3 }genes, respectively. The assembly of collagen VI occurs as a highly orchestrated, multistep process: the three alpha chains intertwine into a triple-helical, heterotrimeric monomer, which subsequently dimerizes and assembles into tetramers. Tetramers are secreted into the extracellular space where they associate to form a beaded microfibrillar network. Pathogenic variants in any of the three \textit{COL6} genes disrupt collagen VI assembly and function, leading to a spectrum of COL6-related muscular dystrophies, ranging from the mild Bethlem muscular dystrophy (BM), via an intermediate phenotype, to the severe Ullrich congenital muscular dystrophy (UCMD). \cite{bonnemann2011collagen}

A diagnosis of collagen VI-related dystrophies (COL6-RD) can be suspected in an individual with characteristic clinical features, muscle imaging features, muscle immunohistochemical features, and collagen VI abnormalities in dermal fibroblast cultures. \cite{foley2021collagen} This initial subjective evaluation is mostly qualitative, and definitive diagnosis relies on the identification of a pathogenic variant in one of the three \textit{COL6} genes via molecular genetic testing. However, diagnosis remains challenging despite advances in next-generation sequencing technologies. The complexity of variant identification has increased with the discovery of more intronic variants that retain introns at low levels, complicating the interpretation of genetic data (C.B., unpublished data). Furthermore, there are still many variants of unknown significance, adding another layer of difficulty in making an accurate diagnosis. Although some methods for quantification have been developed~\cite{osegui2025collablots}, they do not consider the microfibrillar structure that is crucial in assessing the function of collagen VI; therefore, alternative quantitative methodologies to evaluate collagen VI abnormalities are needed. 

\begin{figure}[hbtp]
    \centering
    \includegraphics[width=1\linewidth]{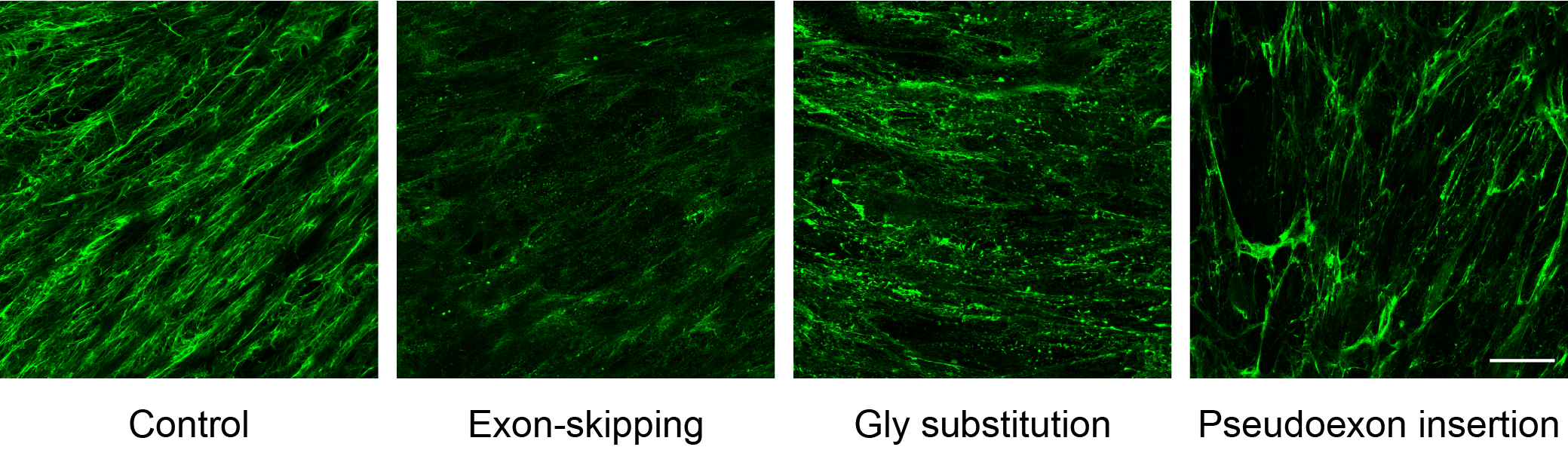}
    \caption{Collagen VI immunofluorescence images from control and patient-derived dermal fibroblast cultures. Each image represents one of the four classes. Scale bar = 50 $\mu$m.}
    \label{fig:col-images}
\end{figure}

COL6-RD is caused either by recessive loss-of-function variants or, more commonly, by \textit{de novo} dominant-negative pathogenic variants. Biallelic recessive variants typically result in loss-of-function or nonsense-mediated RNA decay, and are readily identified by collagen VI immunostaining as the collagen VI signal is reduced or absent. Heterozygous dominant-negative variants, in contrast, can be broadly categorized into three main groups based on their pathogenic mechanism: (i) pathogenic variants causing in-frame exon skipping (\textit{e.g.} splice site variants, in-frame deletions) in any of the three \textit{COL6} genes, (ii) pathogenic variants causing single-amino acid substitutions which disrupt the Gly-X-Y motif of the highly conserved N-terminal triple helical domain of any of the three \textit{COL6 }genes, and (iii) the recurrent deep-intronic pathogenic variant in intron 11 of \textit{COL6A1} gene (c.930+189C$>$T), which creates a cryptic donor splice site that prompts the insertion of a 72-nt in-frame pseudoexon \cite{bonnemann2011collagen, baker2005dominant, brinas2010early, bolduc2019recurrent, cummings2017improving, lampe2008exon, foley2025characterization}. \textit{De novo} dominant-negative pathogenic variants result in the production of structurally abnormal but still assembly-competent alpha-chains. These mutated chains thus retain the ability to assemble into dimers and tetramers but ultimately lead to the deposition of a dysfunctional collagen VI matrix, as revealed by collagen VI immunostaining. Although collagen VI immunostaining of patient-derived fibroblast cultures can detect collagen VI matrix abnormalities, distinguishing among these three groups based on visual inspection remains challenging due to subtle differences in staining patterns (Figure \ref{fig:col-images}).

\begin{figure}[hbtp]
    \centering
    \includegraphics[scale=0.56]{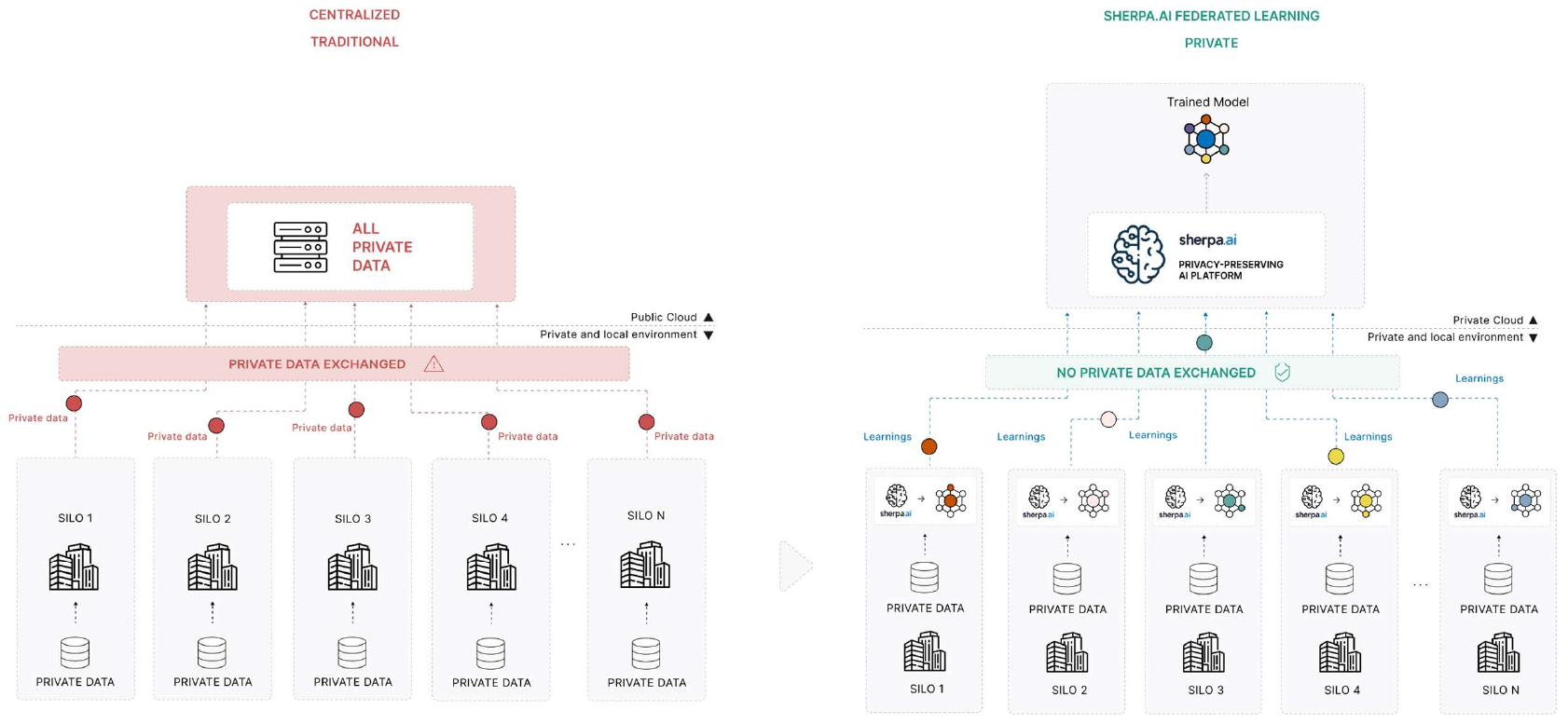}
    \caption{Traditional (Centralized) training (left) compared to Federated training implemented on the Sherpa.ai FL platform (right). In FL, only model updates (not raw images) are shared with a central aggregator, enhancing privacy and regulatory compliance.}
    \label{fig: cl-fl-collection}
\end{figure}

Machine Learning (ML) has become a powerful tool in medical diagnosis and treatment, enabling the identification of complex patterns within biomedical data. However, its successful application in rare diseases research, such as studies involving COL6-RD, is hindered by the limited availability of large and diverse datasets necessary for training accurate and generalizable models. Traditional approaches to data sharing, such as centralized aggregation (Figure \ref{fig: cl-fl-collection} Left), consolidate data from multiple sources onto a single server. While this can increase dataset size and heterogeneity, it raises significant concerns regarding privacy, regulatory compliance, and logistical challenges, including compliance with the General Data Protection Regulation (GDPR) \cite{gdpr_eurlex}, data transmission costs, and restrictions imposed by data localization laws.

Federated Learning (FL) offers a promising alternative by enabling the training of ML models across distributed institutions (nodes) without sharing raw data (Figure \ref{fig: cl-fl-collection}, Right). In FL, each node keeps its data locally and only shares model updates with a central aggregator, which integrates these updates to improve a global model. \cite{mcmahan2017communication}. This approach enhances data privacy, facilitates regulatory compliance, reduces bandwidth requirements, and allows for the inclusion of diverse, geographically distributed datasets. Despite these advantages, FL still faces several challenges, such as security vulnerabilities, increased communication overhead, and the presence of non-independent and identically distributed (non-IID) data across nodes~\cite{liu2022distributed}. In real-world scenarios, non-IID data distributions are common due to variations in patient populations, imaging equipment, and data acquisition protocols. Such heterogeneity can hinder model convergence, reduce performance, and introduce instability or bias during training ~\cite{g2024noniiddatafederatedlearning,lu2024federated}. To address these issues, recent research has explored various strategies, including personalization techniques, adaptive aggregation schemes, and robust optimization methods~\cite{tzortzis2025towards}, aimed at improving model performance under non-IID conditions. 




This study presents a novel multi-institutional FL effort for diagnosing COL6-RD using collagen VI immunofluorescence microscopy images from patient-derived fibroblast cultures. The model was trained on data from 90 COL6-RD patients from two geographically distinct sites across two countries. Specifically, this work describes a new approach to develop a model capable of further subclassifying collagen VI immunofluorescence microscopy images from dermal fibroblast cultures into the three primary pathogenic mechanism groups associated with COL6-RD: exon skipping, glycine substitution, and pseudoexon insertion. By leveraging FL, the model gained knowledge from a larger and more diverse dataset, thereby enhancing its diagnostic utility.

\section{Results}

\subsection{Limited generalization of previous models to external cohorts}

One of the standard laboratory techniques used to investigate suspected cases of COL6-RD is the analysis of collagen VI immunofluorescence microscopy images from dermal fibroblast cultures derived from the proband. This method allows qualitative assessment of collagen VI morphology and extracellular deposition by comparing patient samples with those from healthy controls. 

Building on prior single-institution efforts, Bazaga et al.~\cite{bazaga2019convolutional} reported a two-class convolutional neural network (CNN) trained on z-stacked confocal images, while Frías et al.~\cite{frias2025artificial} systematically compared classical and deep learning approaches—finding that pretrained deep features coupled with Support Vector Machine (SVM) outperformed alternatives (see also \cite{hoang2021practical}). Although these results indicate strong within-site performance, both studies were evaluated exclusively on images from a single institution, leaving out-of-distribution robustness and cross-site generalizability untested.

To overcome these limitations, our study utilized two independently acquired image datasets from geographically distinct sites, \textit{i.e.}, the National Institutes of Health (NIH) dataset and the University College London (UCL) dataset. Each dataset was collected using different microscopy systems and imaging parameters, resulting in a heterogeneous and non-IID collection of images that better represents real-world variability.

To assess the generalizability of Bazaga et al.'s model, we first applied it to the NIH dataset, categorizing images into two classes: healthy control and patient. The model performance was assessed using the F1-score and Accuracy metrics (see Section~\ref{sec: metrics} for metric definitions). 





\begin{table}[h]
  \centering
  \begin{tabularx}{\columnwidth}{Y|Y|Y}
    \toprule
    \textbf{Metric} & \textbf{Bazaga et al. dataset} & \textbf{NIH dataset} \\
    \midrule
    Mean F1-score (STD)   & 0.95 (—)        & 0.689 (0.07) \\
    Mean Accuracy (STD)   & 0.95 (—)        & 0.735 (0.07) \\
    \bottomrule
  \end{tabularx}
  \vspace{10pt}
  \caption{Bazaga et al. model's performance comparison using different datasets.}
  \label{tab: Bazaga_model_comp}
\end{table}

\begin{figure}[hbtp]
    \centering
    \begin{subfigure}{0.48\linewidth}
        \centering
        \includegraphics[width=\linewidth]{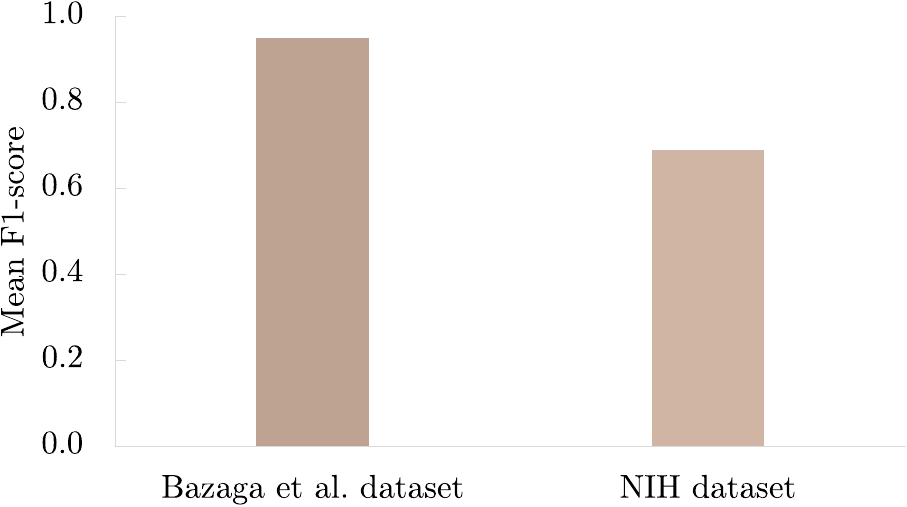}
        \caption{F1-score}
        \label{fig:bazaga_our_dataset_f1_score}
    \end{subfigure}\hfill
    \begin{subfigure}{0.48\linewidth}
        \centering
        \includegraphics[width=\linewidth]{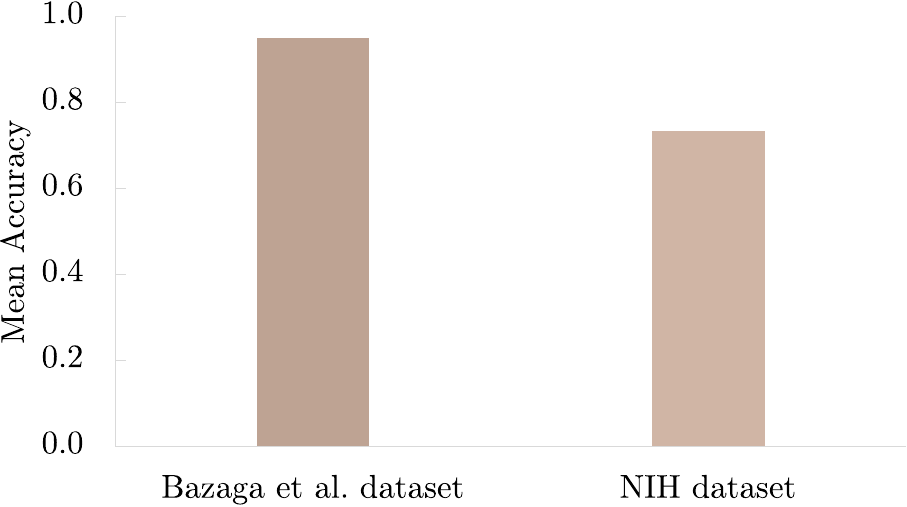}
        \caption{Accuracy}
        \label{fig:bazaga_our_dataset_accuracy}
    \end{subfigure}
    \caption{Mean metrics using Bazaga et al. model with different datasets.}
    \label{fig:bazaga-our-comparison}
\end{figure}

As shown in Table~\ref{tab: Bazaga_model_comp} and Figure~\ref{fig:bazaga-our-comparison}, the model's performance decreases substantially on this external dataset, achieving an F1-score of 0.69 and an Accuracy of 0.74, considerably lower than the originally reported values of 0.95 for both metrics. These results indicate that Bazaga's model has only limited ability to generalize across cohorts with different imaging characteristics and institutional origins.

\subsection{A ML model for subclassifying collagen VI immunofluorescence microscopy images by COL6-RD pathogenic mechanism }
\label{sec: second_model}

Next, we developed an ML model capable of classifying patient collagen VI immunofluorescence images into one of the three pathogenic mechanism groups, thereby performing a four-class classification task (healthy control, glycine substitution, pseudoexon insertion, or exon skipping). The image dataset was defined as: 

\begin{equation}
    \mathbf{X} = \left[ x_{s_{1}}^1, \dots, x_{s_{i}}^{m} \right]^{T}
\end{equation}
 where each element $x_{s_{i}}$ represents a collection of $m$ microscopy images from site $s_{i
} $. The corresponding label vectors are denoted as:  $\mathbf{Y} = \left[ {y_{1}, ..., y_{i}}\right]$ where each $y_{i}\in\left\{1,2,3,4\right\}$, corresponding to the classes: $c = \{\text{control}, \text{glycine substitution}, \text{pseudoexon insertion}, \text{exon skipping}\}$. 

For the classification task, we employed an EfficientNet-B0 neural network (NN) architecture~\cite{tan2019efficientnet}, pretrained on the ImageNet dataset~\cite{imagenet}, as a feature extractor. Feature vectors were generated using Global Average Pooling~\cite{hsiao2019filter}. These features were then classified using a simple NN composed of two dense layers and a dropout layer.
To increase data variability and improve model robustness, data augmentation was applied to the training set.





\begin{table}[h]
  \centering
  \small
  \renewcommand{\arraystretch}{1.3}
  \begin{tabularx}{\columnwidth}{Y|Y|Y}
    \toprule
    \textbf{Metric} & \textbf{Bazaga et al. (2 classes)} & \textbf{Sherpa.ai (converted to 2 classes)} \\
    \midrule
    Mean F1-score (STD)   & 0.689 (0.07) & 0.94 (0.06) \\
    Mean Accuracy (STD)   & 0.735 (0.07) & 0.92 (0.06) \\
    \bottomrule
  \end{tabularx}
  \vspace{10pt}
  \caption{Classification performance comparison between the Bazaga et al. model and our approach for 10-fold cross-validation (NIH local model).}
  \label{tab:table_bazaga_our_comparison}
\end{table}

\begin{figure}[hbtp]
    \centering
    \begin{subfigure}{0.48\linewidth}
        \centering
        \includegraphics[width=\linewidth]{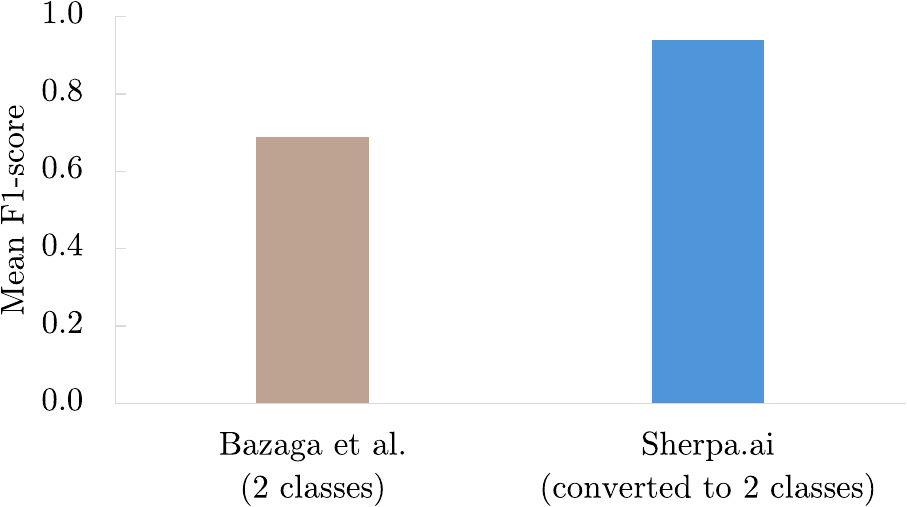}
        \caption{F1-score}
        \label{fig:bazaga_our_model_f1_score}
    \end{subfigure}\hfill
    \begin{subfigure}{0.48\linewidth}
        \centering
        \includegraphics[width=\linewidth]{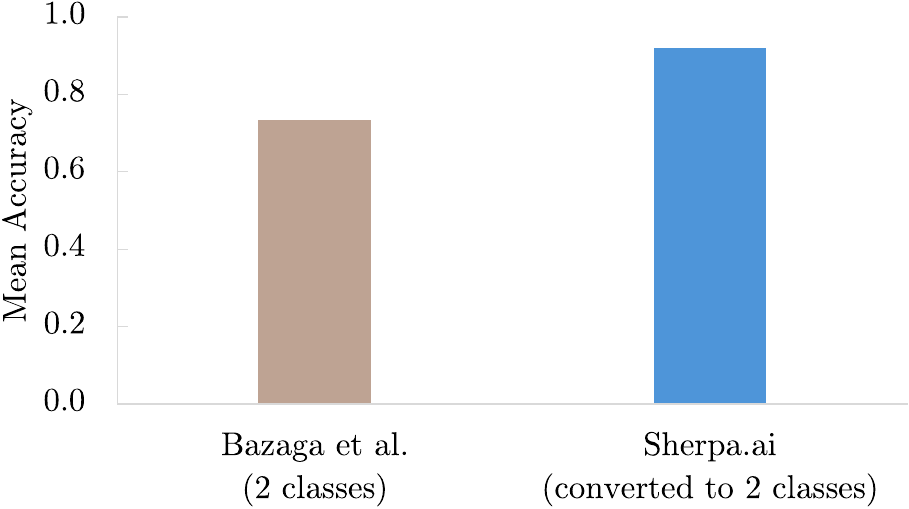}
        \caption{Accuracy}
        \label{fig:bazaga_our_model_accuracy}
    \end{subfigure}
    \caption{Mean metrics comparison between the Bazaga et al. model and our approach for 10-fold cross-validation (NIH local model).}
    \label{fig:bazaga-our-model}
\end{figure}

We assessed the model's performance locally in a non-federated configuration at the NIH site using F1-score and Accuracy as evaluation metrics. The model classified images into four categories: control or one of the three pathogenic mechanism groups. For comparison with the approach described by Bazaga et al., we reformulated this four-class problem into a binary classification task, where all pathogenic groups were combined into a single positive class and the control group was retained as the negative class. Under this formulation, the proposed model achieved an F1-score of 0.94 and an Accuracy of 0.92, demonstrating strong performance in distinguishing patients from controls (see Table~\ref{tab:table_bazaga_our_comparison} and Figure~\ref{fig:bazaga-our-model}).

\subsection{FL enables enhanced pathogenic mechanism-based diagnosis of COL6-RD} 
\label{sec: fed_model}

\begin{figure}
    \centering
    \includegraphics[width=0.46\linewidth]{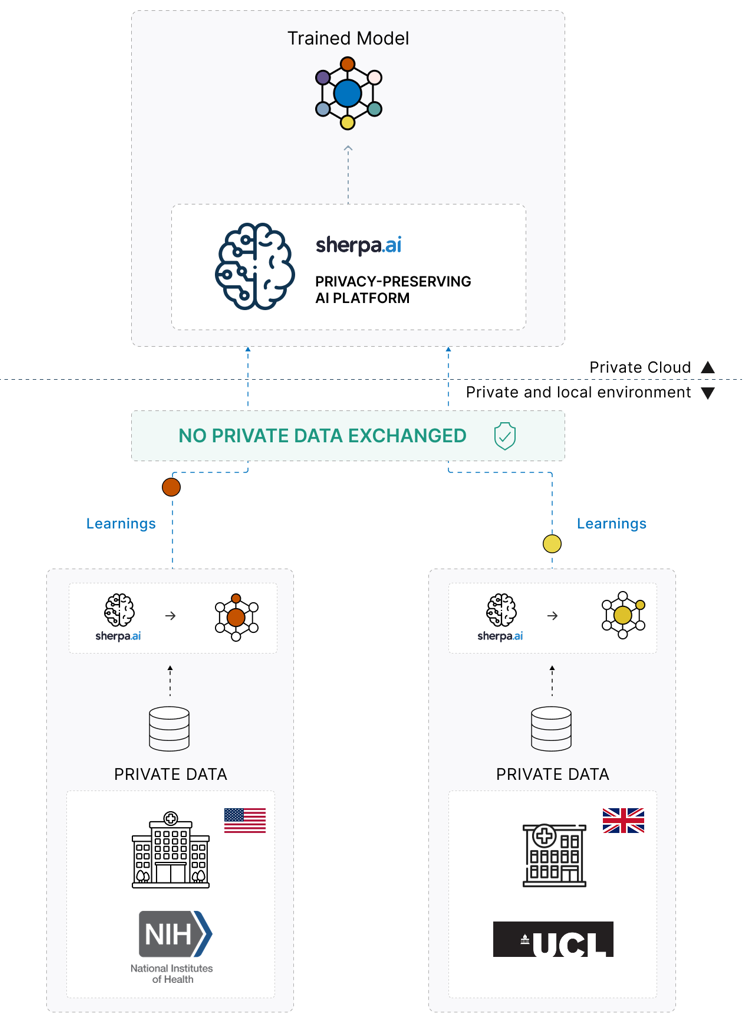}
    \caption{Federated training implemented on the Sherpa.ai FL platform, including the nodes from NIH and UCL.}
    \label{fig:nih-ucl}
\end{figure}

Given the limited data availability, typical of rare diseases such as COL6-RD, we trained our model using FL, which enables multi-institutional training without sharing raw data using the NIH and UCL datasets as nodes (see Figure~\ref{fig:nih-ucl}).

We employed Horizontal FL, a setup where both participating institutions share a common feature space but retain different patient datasets. Each institution (node) trained the local model on its private dataset. These locally learned model parameters were then aggregated using Federated Averaging (FedAvg)~\cite{mcmahan2017communication}. 

Each dataset remains locally stored at its respective institution. Due to differences in imaging protocols, equipment, and preprocessing pipelines, data across sites may vary in quality, resolution, and distribution. To ensure effective model training in this heterogeneous FL setting, we made the following assumptions: (i) Complete class coverage: Nodes complement each other, as some classes may be underrepresented at individual sites, ensuring that the global model benefits from a more balanced dataset. (ii) Standardized image format: All images are resized to a fixed resolution of 256 x 256 pixels and normalized to ensure compatibility with the convolutional neural network architecture. (iii) Domain heterogeneity: We account for inter-site variability in imaging modalities, equipment, patient demographics, and label distributions, introducing data drift typical in real-world, multi-center deployments. This exemplifies a heterogeneous FL scenario. (iv) Label reliability: Labels are annotated by domain experts and assumed to be reliable for supervised training. However, we acknowledge the possible presence of minor to moderate label noise.

For the federated classification task, we utilized the same EfficientNet-B0 feature extractor with global average pooling, as described in Section~\ref{sec: second_model}. To evaluate model performance, we assembled a \textit{hold-out} set of 24 images, 20 from the NIH dataset and 4 from the UCL dataset, the latter limited by data availability. Ten images were selected based on expert criteria, while the remaining images were selected randomly to ensure a balanced class distribution. All splits were performed at the patient level (no patient’s images appear in both training and test).

\begin{figure}[hbtp]
    \centering
    \begin{subfigure}{0.48\linewidth}
        \centering
        \includegraphics[width=\linewidth]{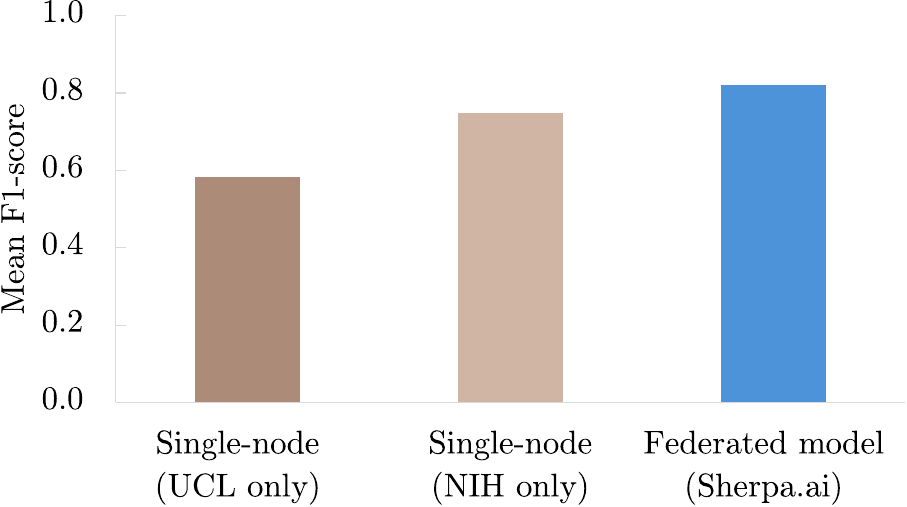}
        \caption{F1-score}
        \label{fig:comparison_single_fl_f1_score}
    \end{subfigure}\hfill
    \begin{subfigure}{0.48\linewidth}
        \centering
        \includegraphics[width=\linewidth]{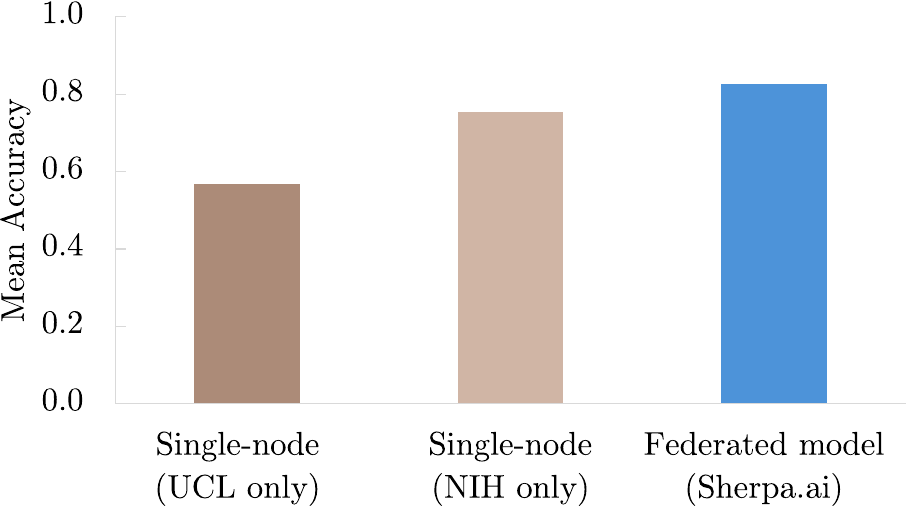}
        \caption{Accuracy}
        \label{fig:comparison_single_fl_accuracy}
    \end{subfigure}
    \caption{Mean metrics across ten random trials for single-node (NIH-UCL) and federated model.}
    \label{fig:single-fl-comparison}
\end{figure}

We used the F1-score as the primary evaluation metric, given its robustness to class imbalance. Accuracy was also reported, consistent with previous FL studies. The model was trained and evaluated under two scenarios: (i) at the single-node level (using only individual datasets, we refer to it in further as the \textit{single-node} model) and (ii) in the federated setting (see Table \ref{tab: table_federated_test} and Figure~\ref{fig:single-fl-comparison}).

\begin{table}[h]
  \centering
  \small
  \renewcommand{\arraystretch}{1.3}
  \begin{tabularx}{\columnwidth}{Y|Y|Y|Y}
    \toprule
    \textbf{Metric} & \textbf{Single-node (NIH only)} & \textbf{Single-node (UCL only)} & \textbf{Federated model (Sherpa.ai)} \\
    \midrule
    Mean F1-score (STD)   & 0.747 (0.024) & 0.582 (0.037) & 0.820 (0.032) \\
    Mean Accuracy (STD)   & 0.754 (0.022) & 0.567 (0.038) & 0.825 (0.031) \\
    \bottomrule
  \end{tabularx}
  \vspace{10pt}
  \caption{Classification performance for UCL and NIH over the test subset — federated and local baseline models. Results shown as means and standard deviations over 10 runs.}
  \label{tab: table_federated_test}
\end{table}

As summarized in Table \ref{tab: table_federated_test}, federated training yielded a mean F1-score of \(0.820 \pm 0.032\) and Accuracy of \(0.825 \pm 0.031\), outperforming both single-node models (NIH-only: F1-score \(0.747 \pm 0.024\), Accuracy \(0.754 \pm 0.022\); UCL-only: F1-score \(0.582 \pm 0.037\), Accuracy \(0.567 \pm 0.038\)). In absolute terms, FL improved F1-score by \(+0.073\) over NIH and \(+0.238\) over UCL (\(\approx +9.8\%\) and \(\approx +40.9\%\) relative, respectively), with corresponding Accuracy gains of \(+0.071\) (\(\approx +9.4\%\)) and \(+0.258\) (\(\approx +45.5\%\)). 

\begin{figure}[hbtp]
    \centering
    \includegraphics[width=0.9\linewidth]{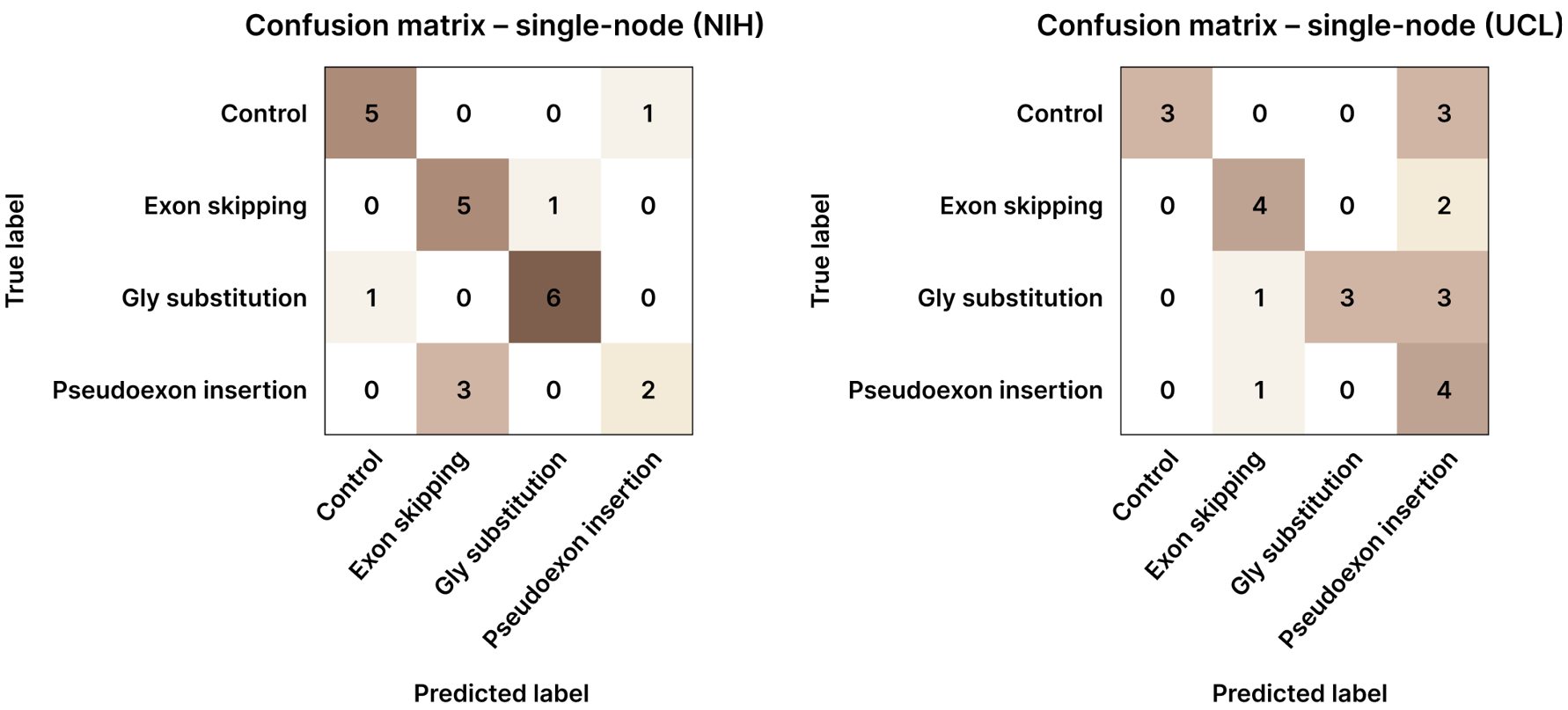}
    \caption{Confusion matrices of the model trained with the NIH dataset (Left) or the UCL dataset (Right), evaluated on the test dataset (\textit{i.e}., before participating in FL). Cell counts reflect images.}
    \label{fig:conf matrix-single-node}
\end{figure}

Figure \ref{fig:conf matrix-single-node} shows confusion matrices for single-node performance: (Left) for the NIH dataset and (Right) for the UCL dataset. The first row in each matrix reflects the classification of control images, with the remaining rows corresponding to patient subgroups. In the NIH dataset (Figure \ref{fig:conf matrix-single-node} Left), 18 out of 24 test images were correctly classified. In contrast, performance on the UCL dataset (Figure \ref{fig:conf matrix-single-node} Right) was lower, with 14 correct classifications and 3 false positives in the control class. 


\begin{figure}[hbtp]
    \centering
    \includegraphics[width=0.58\linewidth]{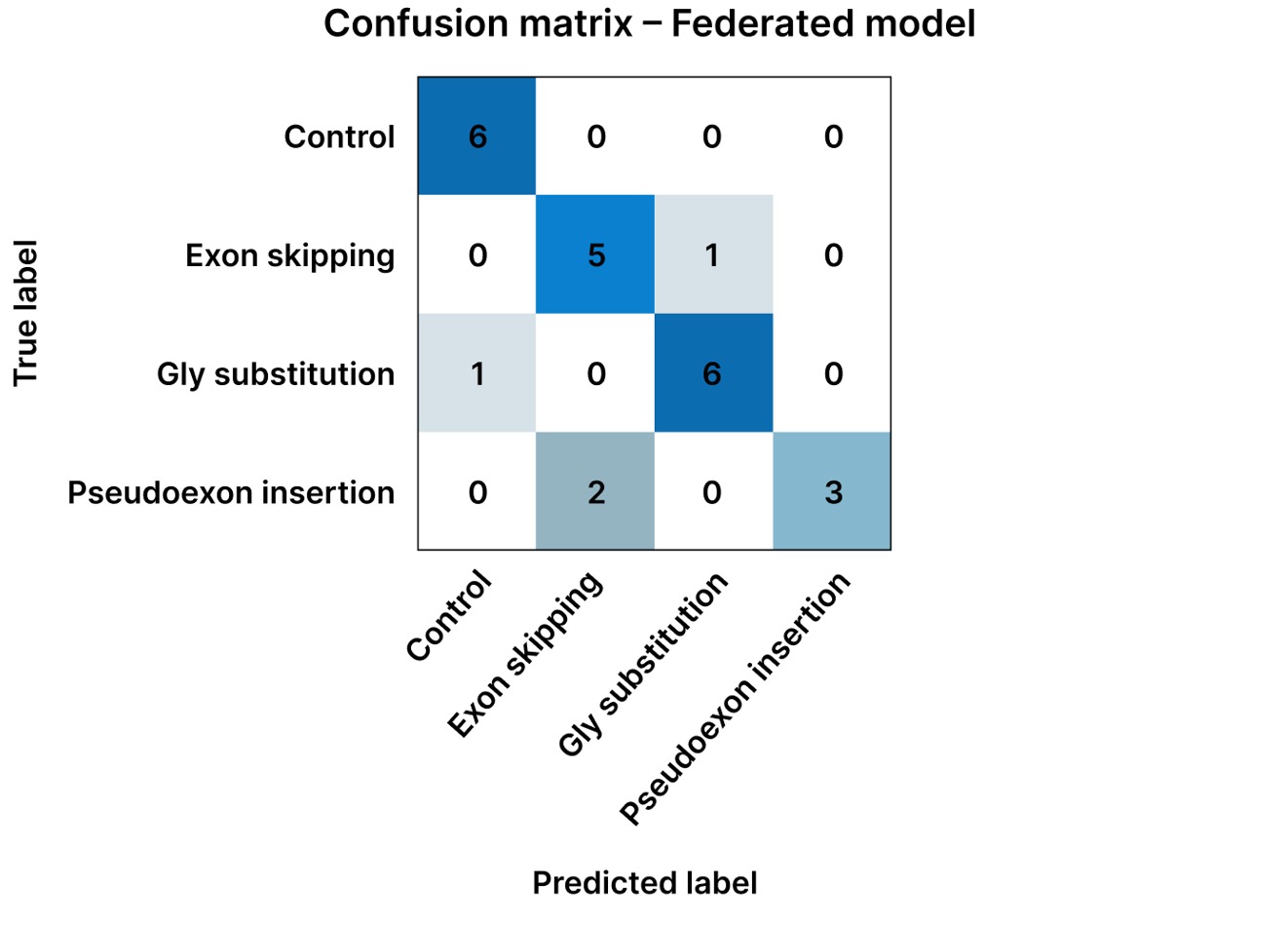}
    \caption{Confusion matrix of the FL model trained with both NIH- and UCL datasets, evaluated on the test dataset (\textit{i.e.}, after participating in FL). Cell counts reflect images.}
    \label{fig:conf matrix-fed}
\end{figure}

Figure~\ref{fig:conf matrix-fed} presents the confusion matrix of the federated model, which outperforms both the NIH single-node model and the UCL single-node results (Figure~\ref{fig:conf matrix-single-node}). These results demonstrate the benefits of federated collaboration, highlighting improved generalization and robustness across institutions.

The confusion matrices in Figures~\ref{fig:conf matrix-single-node} and~\ref{fig:conf matrix-fed} further illustrate the effects highlighted in Table~\ref{tab: table_federated_test}
: while the NIH model maintained stable performance, the UCL site, constrained by a smaller and more heterogeneous dataset, showed marked reductions in misclassifications, particularly fewer false positives in the control class after federation, highlighting improved cross-site generalization from collaborative training.




\section{Discussion}

In this study, we report a novel real-world application of FL for COL6-RD, using data from 90 patients across two geographically distinct sites in two countries. Our objective was to develop a robust, generalizable ML model capable of accurately diagnosing COL6-RD and classifying its underlying pathogenic mechanisms through the analysis of collagen VI matrix immunofluorescence images. In contrast to the public model developed by Bazaga et al.~\cite{bazaga2019convolutional}, which was limited to distinguishing between control and patient images, our approach further stratifies patient images into three distinct pathogenic subgroups. By leveraging the Sherpa.ai privacy-preserving FL platform, we (i) accessed a larger and more diverse dataset, crucial in the context of rare diseases, and (ii) trained our model in a decentralized manner, thereby improving its generalizability to out-of-sample data.

Although previous approaches constituted a substantial improvement on collagen VI matrix assessment within institution baselines, they both left cross-site robustness untested. Bazaga et al.~\cite{bazaga2019convolutional} trained a two-class CNN on z-stacked confocal microscopy images with integrated data augmentation, reporting an Accuracy and F1-score of 0.95, and Frías et al.~\cite{frias2025artificial} showed that pretrained deep features with a linear SVM achieved a biopsy-level AUC of 0.996 (see also \cite{hoang2021practical}). However, these single-institution studies present several limitations that might restrict their broader applicability: (i) limited sample sizes inherent to COL6-RD, yielding small training sets despite data augmentation; (ii) restricted data diversity, as all images originated from a single institution, potentially introducing dataset bias and limiting the model's generalizability to external cohorts; and (iii) imaging system dependency. The dataset was acquired using a specific confocal microscopy system with fixed parameters. While confocal microscopy is widely used, not all laboratories have access to the same equipment or standardized imaging protocols. Moreover, many laboratories primarily rely on wide-field microscopy for rapid, qualitative evaluation of collagen VI matrix, which may further affect the model's performance when applied across diverse settings. These constraints motivated our federated design, which explicitly targets generalization across non-IID, privacy-restricted datasets from different sites.

Over the past decade, ML has shown considerable promise in rare disease research, particularly in the study of rare neuromuscular disorders~\cite{bazaga2019convolutional,chen2025diagnosis,huysmans2023automated,liao2021deep,mastropietro2024classification,schmitt2025deep,verdu2025myo,yang2021deep}. However, the development of effective ML models is often hindered by the limited availability of high-quality patient data, an inherent limitation in rare diseases. This limitation highlights the importance of multi-institutional collaboration and data sharing to aggregate sufficient training data. However, such efforts introduce critical concerns about patient privacy and data security. FL addresses this challenge by enabling collaborative model training without requiring the exchange of raw patient data. Our study demonstrates that FL can facilitate the development of robust and generalizable ML models while preserving data privacy, even in highly data-constrained settings.

The performance of FL systems is highly influenced by the characteristics of the participating institutions. For rare diseases, where data scarcity is pronounced, site selection is limited, and heterogeneity in data quality, quantity, and label distribution across institutions is often unavoidable. In our case, significant non-IID data were present across institutions. While such variability can introduce biases and reduce global model performance, it also reflects the real-world constraints of rare diseases research. Therefore, we incorporated all available data sources in the federation, accepting the resulting heterogeneity as a necessary compromise to maximize data utilization. This highlights the need for FL frameworks that are robust to domain variability and can accommodate the complexity of low-data systems.

Previous studies have investigated FL in data-constrained scenarios, particularly within medical imaging applications~\cite{guan2024federated}. To address this limitation and improve model performance, various strategies have been used, including data augmentation, transfer learning, and feature extraction using pretrained models.~\cite{sheller2020federated,sheller2019multi}. In our study, we adopted a feature extraction approach using models pretrained on the ImageNet dataset, a widely used resource comprising over 14 million labeled images. Transfer learning from ImageNet enabled us to train efficient models with fewer parameters, an approach well-suited to the FL paradigm and small, specialized datasets.

Somewhat unexpectedly, the resulting model also accurately classified collagen VI immunofluorescence patient images according to their underlying pathogenic mechanism. Each institution participating in the FL collaboration benefited differently: For the NIH, the primary benefit was increased model robustness. Although the model weights changed substantially during FL training, the confusion matrices before and after FL (Figures \ref{fig:conf matrix-single-node} and \ref{fig:conf matrix-fed}, respectively) remained very similar. This consistency suggests that the model retained stable performance despite significant parameter updates. At UCL, we observed substantial improvements in predictive performance, particularly in the accurate identification of control images, which the local model had previously misclassified.

Despite these advances, our study has several limitations. First, the dataset was small, reflecting the challenges inherent in rare diseases research. Second, only two sites participated in the FL collaboration, which constrained the diversity and volume of training data. Expanding the federation to include more institutions would enhance data volume and diversity, key factors in improving model performance. Third, considerable heterogeneity was present in the imaging data, including differences in acquisition protocols, equipment, and label distributions. Standardizing image acquisition procedures across sites would help reduce this heterogeneity and enhance model performance. Fourth, the integration of additional data modalities such as magnetic resonance imaging (MRI) images, clinical data, or genomic information warrants further investigation to enhance diagnostic accuracy and model interpretability.

Although FL enables privacy-preserving, multi-institutional collaboration, it is not without risks. The distribution of model parameters and updates across participating sites may introduce vulnerabilities to attacks such as membership inference, gradient leakage, and model inversion. \cite{teo2024federated} Ensuring privacy and security in FL remains an active area of research. Moreover, FL requires each participating site to have sufficient computational infrastructure and reliable communication channels. A privacy-preserving FL platform addresses these challenges, offering a user-friendly interface for coordinated and secure model training across multiple sites.

In conclusion, we have demonstrated the feasibility and utility of an FL platform approach to develop a robust and generalizable ML model capable of classifying collagen VI immunofluorescence images according to their underlying pathogenic mechanisms. This work supports the integration of ML into the diagnostic pipeline for COL6-RD, offering potential support for clinicians, geneticists, and researchers. Specifically, the model may (i) aid in interpreting variants of unknown significance, (ii) guide the prioritization of sequencing strategies to identify novel pathogenic variants, thereby reducing diagnostic time and cost, and (iii) support the development and evaluation of targeted therapies by quantifying their impact on the collagen VI matrix.

\begin{figure}[H]
    \centering
    \includegraphics[width=0.7\linewidth]{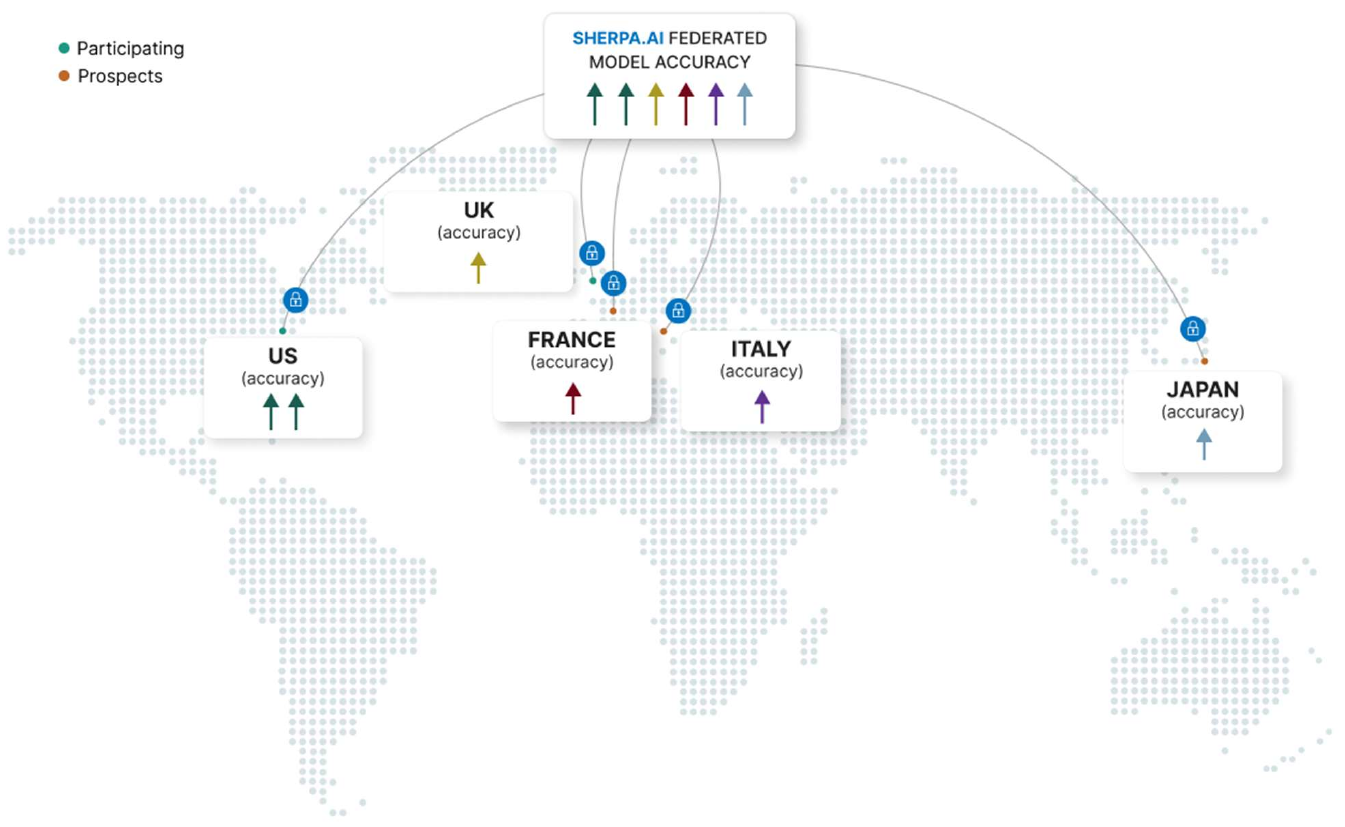}
    \caption{Global map illustrating current and prospective institutions joining the federated learning initiative to enhance model generalizability for collagen VI–related dystrophies.}
    \label{fig:future_world_accuracy_comparison}
\end{figure}

In future work, expanding the federated network to include additional institutions will be essential to enhance the model’s robustness and generalizability across broader patient populations and imaging settings (see Figure~\ref{fig:future_world_accuracy_comparison}). We will prioritize partnerships with international groups and clinical centers that actively study collagen VI–related dystrophies, such as those in France, Italy, and Japan. As participating institutions collect more patient-derived samples, we aim to increase both the quantity and quality of imaging data per node, thereby improving intra-node diversity and overall model performance.

\section{Materials and methods}

\subsection{Data acquisition}
The data used in this study describe patient populations with a confirmed diagnosis of COL6-RD, specifically displaying the collagen VI matrix features, investigated by collagen VI immunofluorescence analysis in dermal fibroblast cultures \cite{jimenez2006comparative}. 
Detailed genetic and demographic characteristics of the study population are provided in Tables~\ref{tab:S1},~\ref{tab:S2}, and ~\ref{tab:S3}.

Dermal fibroblast cultures were established from skin biopsies obtained based on standard operating procedures (12-N-0095, approved by the institutional review board of the National Institute of Neurological Disorders and Stroke (NINDS), and 06/Q0406/33, approved by the institutional review board of the University College London (UCL). Informed consent was obtained from each participant.  

Primary dermal fibroblasts were cultured in Dulbecco’s modified Eagle’s medium (DMEM) supplemented with 10\% fetal bovine serum (FBS) and 1\% penicillin/streptomycin at 37ºC under 5\% CO$_2$. For immunostaining analysis, primary dermal fibroblasts were seeded into 8-chamber tissue culture slides or on collagen-precoated glass coverslips (Corning) in a six-well plate. The following day, the culture medium was replaced with DMEM supplemented with 10\% FBS and 50~$\mu$g/mL of L-ascorbic acid (Wako Chemicals USA, Richmond, VA). 
Every 48-72h, the medium was again replaced and supplemented with L-ascorbic acid. After 3-7 days of L-ascorbic supplementation, cells were fixed with 4\% paraformaldehyde for 10 min at room temperature and matrix immunostaining was performed following the protocol and using antibodies previously described \cite{brull2024optimized}. Alternatively, a collagen VI mouse monoclonal antibody (MAB3303, Chemicon) was also used.  

Images were acquired from multiple microscope systems: a TCS SP5 II system, a Stellaris 8 system (Leica Microsystems, Buffalo Grove, IL), and a Nikon Eclipse Ti microscope (Nikon Instruments, Melville, NY) equipped with an sCMOS pco.edge 4.2 LT camera (Excelitas PCO GmbH, Kelheim, Germany), LSM710 Zeiss Confocal microscope and Leica DMR fluorescence microscope. For confocal images, the pinhole was set to 1 Airy unit, and stack images were acquired using an x40 objective with system-optimized settings. Stacks were then merged. Non-stacked wide-field microscopy images were acquired with x20 or x40 objectives.

\subsection{Data processing}
To develop the model, we assembled two image datasets (\textit{i.e.} NIH and UCL datasets) comprising 331 images from 90 patients collected at two international sites, with varying contributing images across sites (Table~\ref{tab: image_distribution}).

\begin{table}[h]
  \centering
  \small
  \renewcommand{\arraystretch}{1.3}
  \begin{tabularx}{\columnwidth}{Y|Y|Y|Y|Y}
    \toprule
    \textbf{Site ID} & \textbf{Pathogenic mechanism’s group} & \textbf{\# images} & \textbf{\% images} & \textbf{Total \# images} \\
    \midrule
    \multirow{4}{*}{\textbf{NIH}}
      & Control               & 84 & 28.0\% & \multirow{4}{*}{300} \\
      & Exon skipping         & 71 & 23.7\% &  \\
      & Gly substitution      & 94 & 31.3\% &  \\
      & Pseudoexon insertion  & 51 & 17.0\% &  \\
    \midrule
    \multirow{4}{*}{\textbf{UCL}}
      & Control               & 7  & 22.6\% & \multirow{4}{*}{31} \\
      & Exon skipping         & 11 & 35.5\% &  \\
      & Gly substitution      & 5  & 16.1\% &  \\
      & Pseudoexon insertion  & 8  & 25.8\% &  \\
    \bottomrule
  \end{tabularx}
  \vspace{10pt}
  \caption{Image distribution across sites and COL6-RD pathogenic mechanism groups.}
  \label{tab: image_distribution}
\end{table}

For NIH-only experiments reported in Section~\ref{sec: second_model}, we employed 10-fold cross-validation. For federated experiments reported in Section~\ref{sec: fed_model}, evaluation was performed on a fixed hold-out test set of 24 images (20 NIH, 4 UCL), with the remaining images used for training. Given the limited sample size, a data augmentation and preprocessing pipeline was applied to the training set to increase data diversity and improve model robustness. Data augmentation techniques included (i) 45-degree rotations to improve orientation invariance, (ii) horizontal flipping to simulate positional variability, and (iii) brightness adjustments to enhance contrast between collagen VI and background. For brightness augmentation, images were converted to HSV (Hue, Saturation, Value) color space, and the Value (V) channel was scaled by factors 1.25 and 1.5 before reconversion to RGB (Red, Green, Blue).
Subsequently, the images were normalized by scaling the pixel values to the [0, 1] range and resized to 256 x 256 pixels to conform to the model’s input specifications.  
For the validation and test data, no preprocessing was performed besides resizing images to 256 x 256 and normalizing them to the [0, 1] range.

\subsection{ML approach}
\label{subsec: ml-approach}
We formulated the prediction task as a supervised ML problem, aiming to approximate an unknown function that maps inputs to outputs using a finite set of labeled examples. This was framed as an empirical risk minimization problem, with an added regularization term to prevent overfitting. Model training involves optimizing a parameterized function class, such as neural networks or ensemble methods, based on this objective. 

To enable collaborative learning across multiple data holders without requiring data centralization, we extended this framework to a Horizontal FL setting (see the supplementary material). In this setup, each participating node trains a local model on its own data subset and contributes to the global model through iterative aggregation of locally optimized updates. This approach preserves data privacy while aiming to approximate the performance of a fully centralized model.

A detailed mathematical formulation of both the centralized and FL setups, including conditions for optimization and existence of solutions, is provided in the supplementary material (Sections~\ref{supp_mat_sec: ml-approach} and~\ref{supp_mat_sec: fedlearning-hfl-math}).

All experiments were conducted on a machine equipped with 1TB of disk space, an Intel Core i7-7700 4-core CPU at 3.60 GHz, 64 GB of RAM, and running Ubuntu 24.04 as the operating system. We used Python 3.11, TensorFlow 2.14.0, and scikit-learn 1.6.1 for feature extraction and model training.

In the idealized centralized scenario, where all data is pooled in one location (typically prohibited by data protection laws such as the GDPR \cite{gdpr_eurlex}), model performance serves as a benchmark for comparison. Conversely, in the fully private single-node scenario, where neither data nor model parameters are shared, performance typically reflects a lower limit. The federated setting thus aims to achieve performance within this spectrum, striking a balance between utility and privacy.

\subsection{Metrics}
\label{sec: metrics}

In healthcare and medical imaging, performance metrics such as Accuracy, macro-averaged F1-score, and class-specific F1-scores are widely used to evaluate multiclass classification models~\cite{owusu2023imbalanced,salmi2024handling}. While Accuracy provides an overall measure of the proportion of correctly predicted samples across all classes, the F1-score is particularly valuable in the presence of class imbalance, as it balances precision and recall for each class.

For a given class $c$, precision and recall are defined as:  

\begin{equation}
    \text{Precision}_c = \frac{\text{TP}_c}{\text{TP}_c + \text{FP}_c}, \quad \text{Recall}_c = \frac{\text{TP}_c}{\text{TP}_c + \text{FN}_c},
\end{equation}

where $\text{TP}_c$, $\text{FP}_c$, and $\text{FN}_c$ denote the number of true positives, false positives, and false negatives for class $c$, respectively. The F1-score for class $c$ is the harmonic mean of its precision and recall:  

\begin{equation}
    \text{F1}_c = 2 \cdot \frac{\text{Precision}_c \cdot \text{Recall}_c}{\text{Precision}_c + \text{Recall}_c}.
\end{equation}

To provide a single performance indicator over all $K$ classes, the macro-averaged F1-score is computed as the arithmetic mean of the class-wise F1-scores:  

\begin{equation}
    \text{F1}_\text{macro} = \frac{1}{K} \sum_{c=1}^{K} \text{F1}_c.
\end{equation}

The overall Accuracy of the model is calculated as:  

\begin{equation}
    \text{Accuracy} = \frac{1}{N} \sum_{i=1}^{N} \mathds{1}(\hat{y}_i = y_i)
\end{equation}

where $N$ is the total number of samples.  

This evaluation framework ensures that performance is not dominated by the majority classes, providing a balanced view of classification quality across all disease phenotypes.

\section{Data availability}

The datasets used in this study are not made available due to restrictions imposed by acquiring sites.  

\section{Code availability}

The code used in this study is not publicly available and cannot be shared due to institutional policies.

\section{Acknowledgements}
This research was supported in part by the Intramural Research Program of the National Institutes of Health (NIH). The contributions of the NIH authors were made as part of their official duties as NIH federal employees, are in compliance with agency policy requirements, and are considered Works of the United States Government. Additional support was provided by Sherpa.ai, by  Muscular Dystrophy UK (22GRO-PG36-0552) to HZ, SA, and FM, and the Dubowitz Neuromuscular Centre at University College London (UCL). The findings and conclusions presented in this paper are those of the authors and do not necessarily reflect the views of the NIH, the U.S. Department of Health and Human Services, Sherpa.ai, or UCL.
















\section{Author contributions}
A.B. contributed conceptualization, data acquisition, methodology, investigation, visualization, and writing (original draft). S.A. contributed data acquisition, methodology, investigation, and writing (review \& editing). V.B. and Y.H. contributed data acquisition, investigation, and writing (review \& editing). A.A., D.J-G, E.Z. contributed investigation and writing (review \& editing). J.d-R. contributed resources, supervision, project administration, and writing (review \& editing). O.S. contributed methodology, software, and writing (original draft). H.Z, F.M. and C.G.B. contributed conceptualization, resources, funding acquisition, and writing (review \& editing). X.U-E. contributed conceptualization, supervision, resources, funding acquisition, and writing (review \& editing).

\bibliography{references} 






\newpage

\section{Supplementary material}
This section describes the study population’s genetic and demographic characteristics, as well as the mathematical formulation of the ML and FL models.



\setcounter{table}{0}                    
\renewcommand{\thetable}{S\arabic{table}}

\appendix 
\renewcommand{\thesection}{\Alph{section}} 

\section{Genetic and demographic characteristics}

\begin{longtblr}[
  caption={Detailed genetic and demographic characteristics of the study population for NIH.},
  label={tab:S1},
]{
  width=\linewidth,
  colspec={c|c|c|c|c|X[2]|X[1.4]|X[1.7]},
  column{6,7,8} = {halign=c, valign=t},
  cell{2-Z}{6} = {cmd=\seqsplit}, 
  rowhead=1,
  row{1} = {font=\bfseries},          
  rowsep=2pt,
  colsep=4pt
}
\toprule
\textbf{Patient ID} & \textbf{\thead{\#\\images}} & \textbf{Sex} & \textbf{\thead{Age at biopsy \\ (Age at lab \\sample transfer)\\ (y.o.)}
} &
\textbf{Gene} & \textbf{Mutation} & \textbf{Protein Change} &
\textbf{Pathogenic mechanism} \\
\midrule

NIH-1 & 16 & F & 39 & COL6A1 & c.877G{$>$}A & G293R & Gly substitution \\
NIH-2 & 6 & M & (11) & COL6A1 & c.1056+1G{$>$}A & Ex14skip & Exon skipping \\
NIH-3 & 8 & F & 17 & COL6A1 & c.930+189C{$>$}T & Intron11 pseudoexon & Pseudoexon insertion \\
NIH-4 & 4 & M & 16 & COL6A1 & c.868G{$>$}A & G290R & Gly substitution \\
NIH-5 & 5 & F & (11) & COL6A3 & c.6210+1G{$>$}A & Ex16skip & Exon skipping \\
NIH-7 & 15 & M & 18 & COL6A1 & c.877G{$>$}A & G293R & Gly substitution \\
NIH-8 & 15 & M & (7) & COL6A1 & c.877G{$>$}A & G293R & Gly substitution \\
NIH-9 & 39 & N/A & N/A & Control & N/A & N/A & Control \\
NIH-10 & 16 & M & N/A & Control & N/A & N/A & Control \\
NIH-11 & 6 & F & 30 & Control & N/A & N/A & Control \\
NIH-13 & 4 & M & 3 & Control & N/A & N/A & Control \\
NIH-14 & 7 & M & (2) & COL6A2 & c.812G{$>$}A & G271D & Gly substitution \\
NIH-15 & 7 & F & (10) & COL6A1 & c.930+189C{$>$}T & Intron11 pseudoexon & Pseudoexon insertion \\
NIH-16 & 6 & M & 12 & COL6A1 & c.930+189C{$>$}T & Intron11 pseudoexon & Pseudoexon insertion \\
NIH-17 & 6 & F & 18 & COL6A1 & c.930+189C{$>$}T & Intron11 pseudoexon & Pseudoexon insertion \\
NIH-18 & 7 & F & 8 & COL6A1 & c.930+189C{$>$}T & Intron11 pseudoexon & Pseudoexon insertion \\
NIH-19 & 6 & F & 24 & Control & N/A & N/A & Control \\
NIH-20 & 7 & F & (9) & Control & N/A & N/A & Control \\
NIH-21 & 6 & N/A & N/A & Control & N/A & N/A & Control \\
NIH-22 & 8 & F & (3) & COL6A1 & c.930+189C{$>$}T & Intron11 pseudoexon & Pseudoexon insertion \\
NIH-23 & 6 & F & (5) & COL6A1 & c.904-3T{$>$}G & Ex11skip & Exon skipping \\
NIH-24 & 6 & F & 5 & COL6A1 & c.1056+1G{$>$}A & Ex14skip & Exon skipping \\
NIH-25 & 6 & F & (22) & COL6A3 & c.6210+1G{$>$}A & Ex16skip & Exon skipping \\
NIH-26 & 1 & M & 9 & COL6A3 & c.6309+3A{$>$}G & Ex18skip & Exon skipping \\
NIH-27 & 2 & M & 15 & COL6A1 & c.930+189C{$>$}T & Intron11 pseudoexon & Pseudoexon insertion \\
NIH-28 & 1 & M & N/A & COL6A1 & N/A & Ex14skip & Exon skipping \\
NIH-29 & 1 & M & N/A & COL6A3 & c.6156+1G{$>$}A & Ex15skip & Exon skipping \\
NIH-30 & 1 & N/A & N/A & COL6A1 & c.850G{$>$}A & G284R & Gly substitution \\
NIH-31 & 1 & F & (35) & COL6A1 & c.1056+1G{$>$}A & Ex14skip & Exon skipping \\
NIH-32 & 2 & F & (12) & COL6A1 & c.815G{$>$}A & G272D & Gly substitution \\
NIH-33 & 2 & M & (9) & COL6A3 & c.6210+5G{$>$}A & Ex16skip & Exon skipping \\
NIH-34 & 2 & M & (3) & COL6A1 & c.904-10G{$>$}A & Ex11skip & Exon skipping \\
NIH-35 & 1 & F & (40) & COL6A1 & c.859G{$>$}A & G287R & Gly substitution \\
NIH-36 & 1 & M & (15) & COL6A1 & c.850G{$>$}A & G284R & Gly substitution \\
NIH-37 & 2 & F & 26 & COL6A3 & c.6229G{$>$}C & G2077R & Gly substitution \\
NIH-38 & 2 & F & (3) & COL6A1 & c.904-1G{$>$}C & Ex11skip & Exon skipping \\
NIH-39 & 2 & M & (2) & COL6A3 & c.6210+1G{$>$}A & Ex16skip & Exon skipping \\
NIH-40 & 2 & M & (14) & COL6A1 & c.805-2A{$>$}G & Ex9skip & Exon skipping \\
NIH-41 & 2 & F & (3) & COL6A1 & c.842G{$>$}A & G281E & Gly substitution \\
NIH-42 & 2 & M & (8) & COL6A2 & c.1333-1G{$>$}A & Ex16skip & Exon skipping \\
NIH-43 & 2 & M & (7) & COL6A3 & c.6210+5insA & Ex16skip & Exon skipping \\
NIH-44 & 2 & M & 3 & COL6A3 & c.6156+1G{$>$}A & Ex15skip & Exon skipping \\
NIH-45 & 2 & F & 22 & COL6A1 & c.930+189C{$>$}T & Intron 11 pseudoexon & Pseudoexon insertion \\
NIH-46 & 2 & M & (11) & COL6A3 & c.6210+1G{$>$}A & Ex16skip & Exon skipping \\
NIH-47 & 2 & F & (12) & COL6A1 & c.823G{$>$}A & G275R & Gly substitution \\
NIH-48 & 2 & F & (18) & COL6A1 & c.1056+1G{$>$}T & Ex14skip & Exon skipping \\
NIH-49 & 2 & M & 12 & COL6A3 & c.6221G{$>$}A & G2074D & Gly substitution \\
NIH-50 & 2 & M & (14) & COL6A1 & c.1003-2A{$>$}G & Ex14skip & Exon skipping \\
NIH-51 & 2 & F & 30 & COL6A1 & c.805-641\_858+276del & Ex9skip & Exon skipping \\
NIH-52 & 2 & M & (14) & COL6A1 & c.841G{$>$}A & G281R & Gly substitution \\
NIH-53 & 2 & M & (7) & COL6A1 & c.904G{$>$}A & G302R & Gly substitution \\
NIH-54 & 2 & M & (6) & COL6A1 & c.930+189C{$>$}T & Intron 11 pseudoexon & Pseudoexon insertion \\
NIH-56 & 2 & F & (12) & COL6A1 & c.868G{$>$}A & G290R & Gly substitution \\
NIH-57 & 2 & M & 18 & COL6A1 & c.1184G{$>$}T & G395V & Gly substitution \\
NIH-58 & 1 & M & 7 & COL6A1 & c.868G{$>$}A & G290R & Gly substitution \\
NIH-60 & 2 & M & (N/A) & COL6A1 & c.850G{$>$}A & G284R & Gly substitution \\
NIH-61 & 2 & F & 14 & COL6A3 & c.6248G{$>$}A & G2083D & Gly substitution \\
NIH-62 & 2 & F & 2 & COL6A2 & c.901-3\_914del17bp & Ex8skip & Exon skipping \\
NIH-63 & 2 & F & (23) & COL6A1 & c.930+189C{$>$}T & Intron 11 pseudoexon & Pseudoexon insertion \\
NIH-64 & 2 & F & (6) & COL6A3 & c.6157G{$>$}T & G2053C & Gly substitution \\
NIH-65 & 2 & M & (N/A) & COL6A3 & c.6308A{$>$}G & Ex18skip & Exon skipping \\
NIH-66 & 1 & N/A & 7 & COL6A2 & c.855+1G{$>$}A & Ex6skip & Exon skipping \\
NIH-67 & 2 & M & 55 & COL6A1 & c.1056+1G{$>$}A & Ex14skip & Exon skipping \\
NIH-68 & 2 & M & 21 & COL6A1 & c.868G{$>$}A & G290R & Gly substitution \\
NIH-69 & 1 & F & (12) & COL6A1 & c.930+189C{$>$}T & Intron 11 pseudoexon & Pseudoexon insertion \\
NIH-70 & 1 & M & (42) & COL6A1 & N/A & Ex14skip & Exon skipping \\
NIH-72 & 1 & F & 9 & COL6A1 & N/A & Ex14skip & Exon skipping \\
NIH-74 & 1 & M & 3 & COL6A3 & c.6309G{$>$}C & Ex18skip & Exon skipping \\
NIH-75 & 1 & F & 14 & COL6A1 & c.850G{$>$}A & G284R & Gly substitution \\
NIH-76 & 1 & M & 7 & COL6A3 & c.6248G{$>$}A & G2083D & Gly substitution \\
NIH-77 & 2 & M & 0 & COL6A1 & c.850G{$>$}A & G284R & Gly substitution \\
NIH-78 & 1 & M & (N/A) & COL6A2 & c.820G{$>$}A & G274S & Gly substitution \\
NIH-79 & 2 & F & (3) & COL6A2 & c.1162G{$>$}A & G388R & Gly substitution \\
NIH-80 & 2 & M & 3 & COL6A2 & c.954G{$>$}A & Ex9skip & Exon skipping \\
NIH-81 & 2 & N/A & N/A & COL6A1 & c.956A{$>$}G & K319R & Exon skipping \\
\bottomrule
\end{longtblr}

\begin{longtblr}[
  caption={Detailed genetic and demographic characteristics of the study population for UCL.},
  label={tab:S2},
]{
  width=\linewidth,
  colspec={c|c|c|c|c|X[2]|X[1.4]|X[1.7]},
  column{6,7,8} = {halign=c, valign=t},
  cell{2-Z}{6,7} = {cmd=\seqsplit}, 
  rowhead=1,
  row{1} = {font=\bfseries},          
  rowsep=2pt,
  colsep=4pt
}
\toprule
\textbf{Patient ID} & \textbf{\thead{\#\\images}} & \textbf{Sex} & \textbf{\thead{Age at \\ biopsy\\ (y.o.)}} &
\textbf{Gene} & \textbf{Mutation} & \textbf{Protein Change} &
\textbf{Pathogenic mechanism} \\
\midrule
UCL-1 & 2 & F & 23 & COL6A1 & c.930+189C{$>$}T & Intron 11 pseudoexon & Pseudoexon insertion \\
UCL-2 & 2 & M & 3 & COL6A1 & c.930+189C{$>$}T & Intron 11 pseudoexon & Pseudoexon insertion \\
UCL-3 & 3 & M & 7 & COL6A1 & c.930+189C{$>$}T & Intron 11 pseudoexon & Pseudoexon insertion \\
UCL-4 & 1 & M & 6 & COL6A1 & c.930+189C{$>$}T & Intron 11 pseudoexon & Pseudoexon insertion \\
UCL-5 & 4 & M & N/A & COL6A1 & c.1056+1G{$>$}A & Ex14skip & Exon skipping \\
UCL-6 & 1 & M & 12 & COL6A1 & c.1776+1G{$>$}A & Ex27skip & Exon skipping \\
UCL-7 & 1 & F & 22 & COL6A2 & c.2839\_2850del & p.Leu947\_Gly950del & Deletion \\
UCL-8 & 1 & N/A & N/A & COL6A1 & c.428+1G{$>$}A & Ex3skip & Exon skipping \\
UCL-9 & 3 & F & 3 & COL6A3 & c.6309+3A{$>$}G & Ex18 skip & Exon skipping \\
UCL-10 & 1 & M & N/A & COL6A2 & c.2839\_2850del & p.Leu947\_Gly950del & Deletion \\
UCL-11 & 1 & F & 29 & COL6A1 & c.1022G{$>$}T   & G341V & Gly substitution \\
UCL-12 & 2 & M & 33 & COL6A1 & c.877G{$>$}A & G293R & Gly substitution \\
UCL-13 & 1 & N/A & N/A & COL6A1 & c.868G{$>$}A & G290R & Gly substitution \\
UCL-14 & 1 & N/A & N/A & COL6A1 & c.868G{$>$}A & G290R & Gly substitution \\
UCL-15 & 7 & F & 54 & Control & N/A & N/A & Control \\
\bottomrule
\end{longtblr}

\begin{table}[h]
  \caption{Sex distribution and age statistics per site.}
  \centering
  \small
  \renewcommand{\arraystretch}{1.3}
  \setlength{\tabcolsep}{4pt}
  \begin{tabularx}{\columnwidth}{Y|Y|Y|Y|c|Y|Y|Y|c|Y|Y|Y}
    \toprule
    \multirow{2}{*}{\textbf{Site ID}}
      & \multicolumn{3}{c|}{\textbf{SEX}}
      & \multicolumn{4}{c|}{\textbf{AGE: Male}}
      & \multicolumn{4}{c}{\textbf{AGE: Female}} \\
    \cmidrule(lr){2-4}\cmidrule(lr){5-8}\cmidrule(l){9-12}
      & \textbf{M} & \textbf{F} & \textbf{N/A}
      & \textbf{Average} & \textbf{Std.\ Dev.} & \textbf{Min} & \textbf{Max}
      & \textbf{Average} & \textbf{Std.\ Dev.} & \textbf{Min} & \textbf{Max} \\
    \midrule
    \textbf{NIH} & 38 & 32 & 5 & 11.7 & 11.0 & 0 & 55 & 15.5 & 10.7 & 2 & 40 \\
    \textbf{UCL} & 7 & 5 & 3 & 12.2 & 10.8 & 3 & 33 & 19.3 & 9.8 & 3 & 29 \\
    \bottomrule
  \end{tabularx}
  \vspace{10pt}

  \label{tab:S3}
\end{table}

\newpage

\section{ML approach}
\label{supp_mat_sec: ml-approach}
Given an input space $\mathcal X\coloneqq \mathbb{R}^d$ and output set $\mathcal Y\subseteq \mathbb{R}^m$, the goal of Supervised ML is roughly to approximate an unknown function
\begin{equation}\label{f}
	f:\mathcal{X}\longrightarrow \mathcal{Y},
\end{equation}
given a dataset $\mathcal D = \left\{\left(\vec{x}^{\,i}, \vec{y}^{\,i}\right)\right\}_{i=1}^N\subset \mathcal{X}\times \mathcal{Y}$, composed of $N$ known but possibly noisy examples, i.e.
\begin{equation}
	\vec{y}^{\,i}\simeq f(\vec{x}^{\,i}).
\end{equation}

This approximation problem is typically formulated as the minimization of an Empirical Risk (ER) \cite{rumelhart1986learning,goodfellow2016deep} on some training data.

To that purpose (together with preprocessing), the dataset
\begin{equation}
\mathcal{D}=\left\{\left(\vec{x}^{\,i},\vec{y}^{\,i}\right)\right\}_{i=1}^N 
\end{equation}
is firstly split into
\begin{enumerate}
	\item[a)] Training data
	\begin{equation}\label{centralzied_train_dataset}
	\mathcal{D}_{\mbox{\tiny{train}}}=\left\{\left(\vec{x}^{\,i},\vec{y}^{\,i}\right)\right\}_{i\in I_{\mbox{\tiny{train}}}};
	\end{equation}
	\item[b)] Testing data
	\begin{equation}\label{centralzied_test_dataset}
	\mathcal{D}_{\mbox{\tiny{test}}}=\left\{\left(\vec{x}^{\,i},\vec{y}^{\,i}\right)\right\}_{i\in I_{\mbox{\tiny{test}}}};
	\end{equation}
	\item[c)] Validation data
	\begin{equation}\label{}
	\mathcal{D}_{\mbox{\tiny{validation}}}=\left\{\left(\vec{x}^{\,i},\vec{y}^{\,i}\right)\right\}_{i\in I_{\mbox{\tiny{validation}}}},
	\end{equation}
\end{enumerate}
with
\begin{equation}
\left\{1,\dots,N\right\}=I_{\mbox{\tiny{train}}}\bigsqcup I_{\mbox{\tiny{test}}} \bigsqcup I_{\mbox{\tiny{validation}}}.
\end{equation}

At this point, for example, the empirical risk can be defined as
\begin{equation}
	J:\*\Theta\longrightarrow \mathbb{R}
\end{equation}
\begin{equation}\label{eq:functional}
J(\*\theta) \coloneqq \frac{1}{ |I_{\mbox{\tiny{train}}}|}\sum_{i\in I_{\mbox{\tiny{train}}}} \loss\Big(\text{f}_{\*\theta}(\vec{x}^{\,i}), \vec{y}^{\,i}\Big) + \lambda \hspace{0.03 cm} \text{Reg}\left(\*\theta\right),
\end{equation}
where
\begin{itemize}
	\item The parameters space $\*\Theta$ is a Hilbert space on $\mathbb{R}$;
	\item The continuous loss function
	\begin{align*}
	\loss:\mathbb{R}^m\times \mathcal{Y}\longrightarrow \mathbb{R}^+
	\end{align*}
	penalizes the mismatch between the predictions $\text{f}_{\*\theta}(\vec{x}^{\,i})$ and the labels $\vec{y}^{\,i}$;
	\item The \textit{regularization} term $\lambda \text{Reg}\left(\*\theta\right)$ penalizes the model overfitting on training data, the effect of this penalization being modulated by the weighting factor $\lambda>0$ and $\text{Reg}:\*\Theta\longrightarrow \mathbb{R}^+$ being a function (e.g., $\text{Reg}\left(\*\theta\right)=\left\|\*\theta\right\|_{\*\Theta}^2$ the squared Hilbertian norm);
	\item The model
	\begin{equation}
		\text{f}_{\*\theta}:\mathbb{R}^d\longrightarrow \mathbb{R}^m
	\end{equation}
	is a function, belonging to a class
	\begin{align*}
	\mathscr{C}=\big\{\text{f}_{\*\theta} \ | \ \*\theta \in \*\Theta\big\},
	\end{align*}
	$\*\theta$ being the so-called trainable parameters; examples of $\mathscr{C}$ are Deep Neural Networks \cite{goodfellow2016deep}, Random Forests \cite{breiman2001random}, Gradient Boosting Decision Trees \cite{chen2016xgboost}, Transformers \cite{vaswani2017attention}, Large Language Models \cite{naveed2023comprehensive} and Residual Neural Networks (ResNets) \cite{he2016deep}; $\text{f}_{\*\theta}$ is designed to approximate \eqref{f}, for an appropriate choice of the parameters $\*\theta$.
\end{itemize}

In the above context, the ML training is reduced to an \textit{empirical risk minimization problem}, formulated as
\begin{align}\label{opt_pb}
\*\theta^\ast \in \underset{{\*\Theta}}{\mbox{argmin}} J(\*\theta).
\end{align}
\begin{remark}[Existence of solutions]
	Existence of a solution to \eqref{opt_pb} might be analysed by the Direct Method in the Calculus of Variations \cite{dacorogna2007direct}.
	
	For instance, existence holds, assuming
	\begin{itemize}
		\item The parameters space $\*\Theta$ is finite dimensional;
		\item The regularisation weighting parameter $\lambda >0$;
		\item The regularisation function is the Hilbertian norm
		\begin{equation}
			\text{Reg}\left(\*\theta\right)=\left\|\*\theta\right\|_{\*\Theta}^2;
		\end{equation}
		\item For any $\vec{x}\in \mathbb{R}^d$, the function
		\begin{equation}
			\*\theta\mapsto \text{f}_{\*\theta}(\vec{x})
		\end{equation}
		is continuous.
	\end{itemize}
\end{remark}

\begin{remark}[Convexity]
	$J$ might not be convex, even if $\loss$ is convex. Indeed, convexity also depends on
	\begin{equation}
	\*\theta\mapsto \text{f}_{\*\theta}.
	\end{equation}
	
	In case $J$ is not strictly convex, even if a global minimiser exists, its uniqueness is not guaranteed.
\end{remark}

The subscript $\*\theta$ is typically dropped to simplify the notation.

\section{FL and Horizontal FL}
\label{supp_mat_sec: fedlearning-hfl-math}

The collaborative network of the present study spans two countries, with data from two distinct sites. 
In this section, we begin with the setting we defined for ML and extend it to FL.

Consider a dataset $\mathcal{D} = \{(\vec{x}^{i}, \vec{y}^{i})\}^{N}_{i=1} \subset X \times Y$ composed of N known but possible noisy examples and a federation made of P parties $\{F_k\}_{k=1}^P$.

Suppose the examples of $\mathcal{D}$ are distributed among parties, i.e, party $F_{k}$ owns the $k$-th dataset 
\begin{equation}
\end{equation}

where $N = \bigsqcup_{k=1}^{P} I_{k} $ where the symbol $\bigsqcup$ represents a disjoint union. Therefore, in this context, the dataset $\mathcal{D}$ is partitioned as follows
\begin{equation}
    \mathcal{D}_k \coloneqq \{ (\vec{x}^{\,i}, \vec{y}^{\,i}) \mid i \in I_k \}
\label{eq:dataset}
    \mathcal{D} = \bigsqcup_{k=1}^{P} \mathcal{D}_{k}
\end{equation}
, which can be seen as a horizontal slicing. 
The dataset $\mathcal{D}$ is defined as a table where the examples, that is, the images, are its rows.

In this context, our goal is approximating an \textit{unknown function}$f: \mathcal{X} \longrightarrow \mathcal{Y}$ from the $N$ known but possibly noisy examples $\{(\vec{x}^{i}, \vec{y}^{i})\}^{N}_{i=1} \subset X \times Y$ dispersed across parties as in \ref{eq:dataset}.

In a standard Supervised ML scenario, this is often reduced to an \textit{empirical risk minimization}, as described in Section \ref{subsec: ml-approach}.

In this horizontal scenario, the examples $(\vec{x}^{i}, \vec{y}^{i})$ are dispersed across parties, whence no party can fully evaluate the functional defined in \ref{eq:functional}. However, we can rewrite the function as the average of functions that can be handled by each party
\begin{equation}
    \label{eq: fed-func}
    J(\*\theta) = \sum_{k=1}^{P} \lambda_{k}J_{k}(\*\theta)
\end{equation}
where $N_{k} \coloneqq |I_{k}|$  is the number of examples possessed by party $F_{k}$ and the weighting parameter $\lambda_{k} \coloneqq \frac{N_{k}}{N}$ and 
\begin{equation}
    \label{equation:functional-fed}
    J_{k}(\*\theta) \coloneqq \frac{1}{N_{k}} \sum_{i \in I_{k}} \loss\Big(\text{f}_{\*\theta}(\vec{x}^{\,i}), \vec{y}^{\,i}\Big) + \frac{\alpha N}{N_{k}P} ||\*\theta||^{2}_{\*\Theta}
\end{equation}
Note that, if $N_{k} = \frac{N}{P}$ for every $k \in ||P||$, we have $\frac{\alpha N}{N_{k}P} = \alpha$.

Hence, in a federated setting, the minimization \ref{eq:functional} can be reduced to 
\begin{enumerate}[label=\alph*)]
    \item Minimization of $J_{k}$, which is a local task assigned to party $F_{k}$
    \item Aggregation of the local results
\end{enumerate}
The problem of Horizontal FL is to interlace a) and b) to have an optimal trade-off between privacy, predictive performance, and overhead of communications.

\end{document}